\documentclass[runningheads]{llncs}

\usepackage{eccv}

\usepackage{eccvabbrv}

\usepackage{graphicx}
\usepackage{booktabs}

\usepackage[accsupp]{axessibility}  

\usepackage{multirow}
\usepackage{colortbl}
\usepackage{pifont}
\usepackage{adjustbox}
\usepackage{bm}
\usepackage{multirow}
\usepackage{xcolor}
\usepackage{algorithm}
\usepackage{algpseudocode}

\usepackage[pagebackref,breaklinks,colorlinks,citecolor=eccvblue]{hyperref}

\usepackage{orcidlink}

\begin{document}

\title{Conditional Distribution Modelling for Few-Shot Image Synthesis with Diffusion Models} 

\titlerunning{CDM}

\author{Parul Gupta\inst{1}\orcidlink{0000-0002-4379-1573} \and
Munawar Hayat\inst{1}\orcidlink{0000-0002-2706-5985} \and
Abhinav Dhall\inst{2}\orcidlink{0000-0002-2230-1440} \and
Thanh-Toan Do\inst{1}
}

\authorrunning{P.~Gupta et al.}

\institute{Monash University, Clayton VIC 3168, Australia \\
\email{\{parul,munawar.hayat,toan.do\}@monash.edu}\and
Flinders University, Adelaide SA 5042, Australia\\
\email{abhinav.dhall@flinders.edu.au}}

\maketitle
\global\long\def\sidenote#1{\marginpar{\small\emph{{\color{Medium}#1}}}}%

\global\long\def\se{\hat{\text{se}}}%
\global\long\def\interior{\text{int}}%
\global\long\def\boundary{\text{bd}}%
\global\long\def\ML{\textsf{ML}}%
\global\long\def\GML{\mathsf{GML}}%
\global\long\def\HMM{\mathsf{HMM}}%
\global\long\def\support{\text{supp}}%
\global\long\def\new{\text{*}}%
\global\long\def\stir{\text{Stirl}}%
\global\long\def\mA{\mathcal{A}}%
\global\long\def\mB{\mathcal{B}}%
\global\long\def\expect{\mathbb{E}}%
\global\long\def\mF{\mathcal{F}}%
\global\long\def\mK{\mathcal{K}}%
\global\long\def\mH{\mathcal{H}}%
\global\long\def\mX{\mathcal{X}}%
\global\long\def\mZ{\mathcal{Z}}%
\global\long\def\mS{\mathcal{S}}%
\global\long\def\Ical{\mathcal{I}}%
\global\long\def\mT{\mathcal{T}}%
\global\long\def\Pcal{\mathcal{P}}%
\global\long\def\dist{d}%
\global\long\def\HX{\entro\left(X\right)}%
\global\long\def\entropyX{\HX}%
\global\long\def\HY{\entro\left(Y\right)}%
\global\long\def\entropyY{\HY}%
\global\long\def\HXY{\entro\left(X,Y\right)}%
\global\long\def\entropyXY{\HXY}%
\global\long\def\mutualXY{\mutual\left(X;Y\right)}%
\global\long\def\mutinfoXY{\mutualXY}%
\global\long\def\given{\mid}%
\global\long\def\gv{\given}%
\global\long\def\goto{\rightarrow}%
\global\long\def\asgoto{\stackrel{a.s.}{\longrightarrow}}%
\global\long\def\pgoto{\stackrel{p}{\longrightarrow}}%
\global\long\def\dgoto{\stackrel{d}{\longrightarrow}}%
\global\long\def\lik{\mathcal{L}}%
\global\long\def\logll{\mathit{l}}%
\global\long\def\bigcdot{\raisebox{-0.5ex}{\scalebox{1.5}{\ensuremath{\cdot}}}}%
\global\long\def\sig{\textrm{sig}}%
\global\long\def\likelihood{\mathcal{L}}%
\global\long\def\vectorize#1{\mathbf{#1}}%

\global\long\def\vt#1{\mathbf{#1}}%
\global\long\def\gvt#1{\boldsymbol{#1}}%
\global\long\def\idp{\ \bot\negthickspace\negthickspace\bot\ }%
\global\long\def\cdp{\idp}%
\global\long\def\das{}%
\global\long\def\id{\mathbb{I}}%
\global\long\def\idarg#1#2{\id\left\{  #1,#2\right\}  }%
\global\long\def\iid{\stackrel{\text{iid}}{\sim}}%
\global\long\def\bzero{\vt 0}%
\global\long\def\bone{\mathbf{1}}%
\global\long\def\a{\mathrm{a}}%
\global\long\def\ba{\mathbf{a}}%
\global\long\def\b{\mathrm{b}}%
\global\long\def\bb{\mathbf{b}}%
\global\long\def\B{\mathrm{B}}%
\global\long\def\boldm{\boldsymbol{m}}%
\global\long\def\c{\mathrm{c}}%
\global\long\def\C{\mathrm{C}}%
\global\long\def\d{\mathrm{d}}%
\global\long\def\D{\mathrm{D}}%
\global\long\def\N{\mathrm{N}}%
\global\long\def\h{\mathrm{h}}%
\global\long\def\H{\mathrm{H}}%
\global\long\def\bH{\mathbf{H}}%
\global\long\def\K{\mathrm{K}}%
\global\long\def\M{\mathrm{M}}%
\global\long\def\bff{\vt f}%
\global\long\def\bx{\mathbf{\mathbf{x}}}%

\global\long\def\bl{\boldsymbol{l}}%
\global\long\def\s{\mathrm{s}}%
\global\long\def\T{\mathrm{T}}%
\global\long\def\bu{\mathbf{u}}%
\global\long\def\v{\mathrm{v}}%
\global\long\def\bv{\mathbf{v}}%
\global\long\def\bo{\boldsymbol{o}}%
\global\long\def\bh{\mathbf{h}}%
\global\long\def\bs{\boldsymbol{s}}%
\global\long\def\x{\mathrm{x}}%
\global\long\def\bx{\mathbf{x}}%
\global\long\def\bz{\mathbf{z}}%
\global\long\def\hbz{\hat{\bz}}%
\global\long\def\z{\mathrm{z}}%
\global\long\def\y{\mathrm{y}}%
\global\long\def\bxnew{\boldsymbol{y}}%
\global\long\def\bX{\boldsymbol{X}}%
\global\long\def\tbx{\tilde{\bx}}%
\global\long\def\by{\mathbf{y}}%
\global\long\def\bY{\boldsymbol{Y}}%
\global\long\def\bZ{\boldsymbol{Z}}%
\global\long\def\bU{\boldsymbol{U}}%
\global\long\def\bn{\boldsymbol{n}}%
\global\long\def\bV{\boldsymbol{V}}%
\global\long\def\bI{\boldsymbol{I}}%
\global\long\def\J{\mathrm{J}}%
\global\long\def\bJ{\mathbf{J}}%
\global\long\def\w{\mathrm{w}}%
\global\long\def\bw{\vt w}%
\global\long\def\bW{\mathbf{W}}%
\global\long\def\balpha{\gvt{\alpha}}%
\global\long\def\bdelta{\boldsymbol{\delta}}%
\global\long\def\bsigma{\gvt{\sigma}}%
\global\long\def\bbeta{\gvt{\beta}}%
\global\long\def\bmu{\gvt{\mu}}%
\global\long\def\btheta{\boldsymbol{\theta}}%
\global\long\def\blambda{\boldsymbol{\lambda}}%
\global\long\def\bgamma{\boldsymbol{\gamma}}%
\global\long\def\bpsi{\boldsymbol{\psi}}%
\global\long\def\bphi{\boldsymbol{\phi}}%
\global\long\def\bpi{\boldsymbol{\pi}}%
\global\long\def\bomega{\boldsymbol{\omega}}%
\global\long\def\bepsilon{\boldsymbol{\epsilon}}%
\global\long\def\btau{\boldsymbol{\tau}}%
\global\long\def\bxi{\boldsymbol{\xi}}%
\global\long\def\realset{\mathbb{R}}%
\global\long\def\realn{\realset^{n}}%
\global\long\def\integerset{\mathbb{Z}}%
\global\long\def\natset{\integerset}%
\global\long\def\integer{\integerset}%

\global\long\def\natn{\natset^{n}}%
\global\long\def\rational{\mathbb{Q}}%
\global\long\def\rationaln{\rational^{n}}%
\global\long\def\complexset{\mathbb{C}}%
\global\long\def\comp{\complexset}%

\global\long\def\compl#1{#1^{\text{c}}}%
\global\long\def\and{\cap}%
\global\long\def\compn{\comp^{n}}%
\global\long\def\comb#1#2{\left({#1\atop #2}\right) }%
\global\long\def\param{\vt w}%
\global\long\def\Param{\Theta}%
\global\long\def\meanparam{\gvt{\mu}}%
\global\long\def\Meanparam{\mathcal{M}}%
\global\long\def\meanmap{\mathbf{m}}%
\global\long\def\logpart{A}%
\global\long\def\simplex{\Delta}%
\global\long\def\simplexn{\simplex^{n}}%
\global\long\def\dirproc{\text{DP}}%
\global\long\def\ggproc{\text{GG}}%
\global\long\def\DP{\text{DP}}%
\global\long\def\ndp{\text{nDP}}%
\global\long\def\hdp{\text{HDP}}%
\global\long\def\gempdf{\text{GEM}}%
\global\long\def\rfs{\text{RFS}}%
\global\long\def\bernrfs{\text{BernoulliRFS}}%
\global\long\def\poissrfs{\text{PoissonRFS}}%
\global\long\def\grad{\gradient}%
\global\long\def\gradient{\nabla}%
\global\long\def\partdev#1#2{\partialdev{#1}{#2}}%
\global\long\def\partialdev#1#2{\frac{\partial#1}{\partial#2}}%
\global\long\def\partddev#1#2{\partialdevdev{#1}{#2}}%
\global\long\def\partialdevdev#1#2{\frac{\partial^{2}#1}{\partial#2\partial#2^{\top}}}%
\global\long\def\closure{\text{cl}}%
\global\long\def\cpr#1#2{\Pr\left(#1\ |\ #2\right)}%
\global\long\def\var{\text{Var}}%
\global\long\def\Var#1{\text{Var}\left[#1\right]}%
\global\long\def\cov{\text{Cov}}%
\global\long\def\Cov#1{\cov\left[ #1 \right]}%
\global\long\def\COV#1#2{\underset{#2}{\cov}\left[ #1 \right]}%
\global\long\def\corr{\text{Corr}}%
\global\long\def\sst{\text{T}}%
\global\long\def\SST{\sst}%
\global\long\def\ess{\mathbb{E}}%

\global\long\def\Ess#1{\ess\left[#1\right]}%
\global\long\def\fisher{\mathcal{F}}%

\global\long\def\bfield{\mathcal{B}}%
\global\long\def\borel{\mathcal{B}}%
\global\long\def\bernpdf{\text{Bernoulli}}%
\global\long\def\betapdf{\text{Beta}}%
\global\long\def\dirpdf{\text{Dir}}%
\global\long\def\gammapdf{\text{Gamma}}%
\global\long\def\gaussden#1#2{\text{Normal}\left(#1, #2 \right) }%
\global\long\def\gauss{\mathbf{N}}%
\global\long\def\gausspdf#1#2#3{\text{Normal}\left( #1 \lcabra{#2, #3}\right) }%
\global\long\def\multpdf{\text{Mult}}%
\global\long\def\poiss{\text{Pois}}%
\global\long\def\poissonpdf{\text{Poisson}}%
\global\long\def\pgpdf{\text{PG}}%
\global\long\def\wshpdf{\text{Wish}}%
\global\long\def\iwshpdf{\text{InvWish}}%
\global\long\def\nwpdf{\text{NW}}%
\global\long\def\niwpdf{\text{NIW}}%
\global\long\def\studentpdf{\text{Student}}%
\global\long\def\unipdf{\text{Uni}}%
\global\long\def\transp#1{\transpose{#1}}%
\global\long\def\transpose#1{#1^{\mathsf{T}}}%
\global\long\def\mgt{\succ}%
\global\long\def\mge{\succeq}%
\global\long\def\idenmat{\mathbf{I}}%
\global\long\def\trace{\mathrm{tr}}%
\global\long\def\argmax#1{\underset{_{#1}}{\text{argmax}} }%
\global\long\def\argmin#1{\underset{_{#1}}{\text{argmin}\ } }%
\global\long\def\diag{\text{diag}}%
\global\long\def\norm{}%
\global\long\def\spn{\text{span}}%
\global\long\def\vtspace{\mathcal{V}}%
\global\long\def\field{\mathcal{F}}%
\global\long\def\ffield{\mathcal{F}}%
\global\long\def\inner#1#2{\left\langle #1,#2\right\rangle }%
\global\long\def\iprod#1#2{\inner{#1}{#2}}%
\global\long\def\dprod#1#2{#1 \cdot#2}%
\global\long\def\norm#1{\left\Vert #1\right\Vert }%
\global\long\def\entro{\mathbb{H}}%
\global\long\def\entropy{\mathbb{H}}%
\global\long\def\Entro#1{\entro\left[#1\right]}%
\global\long\def\Entropy#1{\Entro{#1}}%
\global\long\def\mutinfo{\mathbb{I}}%
\global\long\def\relH{\mathit{D}}%
\global\long\def\reldiv#1#2{\relH\left(#1||#2\right)}%
\global\long\def\KL{KL}%
\global\long\def\KLdiv#1#2{\KL\left(#1\parallel#2\right)}%
\global\long\def\KLdivergence#1#2{\KL\left(#1\ \parallel\ #2\right)}%
\global\long\def\crossH{\mathcal{C}}%
\global\long\def\crossentropy{\mathcal{C}}%
\global\long\def\crossHxy#1#2{\crossentropy\left(#1\parallel#2\right)}%
\global\long\def\breg{\text{BD}}%
\global\long\def\lcabra#1{\left|#1\right.}%
\global\long\def\lbra#1{\lcabra{#1}}%
\global\long\def\rcabra#1{\left.#1\right|}%
\global\long\def\rbra#1{\rcabra{#1}}%

\begin{abstract}
  Few-shot image synthesis entails generating diverse and realistic images of novel categories using only a few example images. While multiple recent efforts in this direction have achieved impressive results, the existing approaches are dependent only upon the few novel samples available at test time in order to generate new images, which restricts the diversity of the generated images. To overcome this limitation, we propose \textit{Conditional Distribution Modelling (CDM)} -- a framework which effectively utilizes Diffusion models for few-shot image generation. By modelling the distribution of the latent space used to condition a Diffusion process, CDM \textit{leverages the learnt statistics of the training data} to get a better approximation of the unseen class distribution, thereby removing the bias arising due to limited number of few shot samples. Simultaneously, we devise a \textit{novel inversion based optimization strategy} that further improves the approximated unseen class distribution, and ensures the fidelity of the generated samples to the unseen class. The experimental results on four benchmark datasets demonstrate the effectiveness of our proposed CDM for few-shot generation. 
\end{abstract}
\section{Introduction}
\label{sec:intro}
In this paper, we tackle few-shot image synthesis i.e., given only a few samples of a novel category, our goal is to generate diverse and realistic images of this new concept.
Few shot synthesis can be effectively used to generate rarely occurring objects (such as rare species of birds or animals) and help to overcome the class imbalance issue in the naturally occurring datasets \cite{DBLP:journals/corr/abs-1904-05160} by generating more examples of the minority categories. 
However, few-shot synthesis is a challenging task because the state of the art generative models such as GANs \cite{https://doi.org/10.48550/arxiv.1406.2661} and VAEs \cite{https://doi.org/10.48550/arxiv.1312.6114} need a large amount of data to learn any concept. Even with sufficient amount of data, due to their inherent nature (involving adversarial learning) adapting GANs for few samples is unstable to converge \cite{zhao2020differentiable}. On the other hand, the sample quality achieved by VAEs is not as good as GANs even though their training is stable \cite{DBLP:journals/corr/abs-2106-06819}. The scarcity of the available novel examples in few-shot learning makes it difficult to estimate the distribution of the new categories which is disjoint from the training distribution. This often leads to overfitting on the few available test samples and lack of diversity in the generated examples.

Most of the existing few-shot synthesis approaches employ GAN-based architectures with carefully formulated auxiliary loss functions. We can broadly categorize these methods into 3 types: fusion based, optimization based and transformation based. Fusion based approaches \cite{pmlr-v84-bartunov18a,DBLP:journals/corr/abs-2003-03497,DBLP:journals/corr/abs-2008-01999,gu2021lofgan} fuse the latent features of a set of same-class images and then decode the fused feature into a new image to generate sample of the same class. 
However in addition to needing at least two novel examples, the generated images lack diversity, as they are similar to the few available examples and can't cover the entire distribution of the novel class. Optimization based approaches \cite{https://doi.org/10.48550/arxiv.1901.02199,DBLP:journals/corr/abs-2001-00576,zheng2023my} introduce meta-learning, where they first learn a base model and then fine tune it for each novel category to generate new images. But the images generated by these methods are often blurry and of low quality. Transformation based approaches \cite{https://doi.org/10.48550/arxiv.1711.04340,HongDeltaGAN} learn the intra-category transformations in the training data and apply them on the novel samples to generate new examples. Our method is a combination of the transformation and optimization based approaches, having the merits of both, \ie good quality samples belonging to unseen class but with added diversity in the generated images, owing to our modelling the entire distribution of the novel class in the conditional latent space, which yields efficient and meaningful augmentation of the novel class samples (as explained in detail in \cref{section:Conditional_Space_Modelling}).

We use a recently popular class of generative methods called Denoising Diffusion Probabilistic Models \cite{DBLP:journals/corr/abs-2006-11239} for few shot image generation. These models can learn to generate meaningful images belonging to the training distribution from pure noise and enjoy the benefit of a stabilized training process (no adversarial learning) while maintaining the sample quality. They have shown remarkable success in various applications such as text-conditioned image generation \cite{https://doi.org/10.48550/arxiv.2204.06125, https://doi.org/10.48550/arxiv.2205.11487}, image super-resolution \cite{https://doi.org/10.48550/arxiv.2104.07636} and have outperformed GANs in class-conditioned image generation \cite{DBLP:journals/corr/abs-2105-05233}.
Since diffusion models are essentially probability density estimators, their application to few shot generation is non-trivial. Unlike VAEs, their latent space is unstructured -- the latents are just training images disturbed by adding varying levels of Gaussian noise. Therefore, applying the ideas from few-shot GAN-based approaches \cite{HongDeltaGAN, gu2021lofgan} and fusing or transforming the latent representations of diffusion models is not intuitively expected to yield any meaningful results. Diffusion-conditioning mechanisms have been developed for controlling simpler concepts such as low-shot attribute generation \cite{DBLP:journals/corr/abs-2106-06819}, class-conditional synthesis (for seen classes during training) \cite{DBLP:journals/corr/abs-2102-09672}, artistic domain-adaptation \cite{zhu2022few} and test-time adaptation \cite{choi2021ilvr}, but not yet explored to generate entirely novel complex classes, from only a few available examples, using a limited amount of training data.
Our approach successfully generates new samples of unseen concepts by modelling the distribution of the space used to condition the diffusion process. 
In order to do so, we develop \textit{Conditional Distribution Modelling} (see \cref{section:Conditional_Space_Modelling}) where we estimate the probability distribution of samples belonging to each class. This enables us to augment the conditional vectors for novel class by simply sampling from its approximated distribution, whose statistics are borrowed from the closest class in the training set and optimized using the few novel class samples. 
These augmented unseen conditionals when passed through the conditional diffusion process in turn give us a set of diverse and realistic new samples.
While the latent space distribution of training classes can be accurately estimated using the abundant samples available, using only the small number of samples of the never seen concepts at test time results in poor approximation of their distribution statistics. To counter this, for each unseen class, we propose to transfer the distribution statistics from the closest seen classes \cite{yang2021free} to get an initial estimate and optimize them using the available samples.
Thus, the sample diversity in our approach comes from two sources -- firstly, the conditional space modelling helps us generate more conditionals for the diffusion process and secondly the diffusion model learns valid intra-class transformations (from the conditional latent to the target latent).

In summary, our major contributions are: 
\begin{itemize}
\item We propose \textbf{a novel diffusion model-based framework for few-shot image generation} that effectively captures the diversity while maintaining the distinctive characteristics of unseen classes by modelling their distributions in its conditional space.
\item Specifically, instead of relying only on the few-shot examples to generate new samples, we develop a principled approach that leverages the learned statistics from the neighbouring seen classes to approximate the unseen class distribution and \textbf{faithfully capture the unseen class diversity}.
\item Further, we propose \textbf{a novel inversion based optimization to refine the unseen class distribution} 
which in turn ensures the fidelity of the generated samples to the unseen class. 


\end{itemize}
\section{Related Work}
\label{section:ra1_related_work}
The existing literature on few shot image generation can be broadly divided into three categories - fusion-based, optimization-based and transformation-based methods.\\
\textbf{Fusion based approaches} produce new images of unseen classes by \textit{fusing} the latent features of the example images in some manner and passing the fused representation through a decoder, e.g.\ GMN \cite{pmlr-v84-bartunov18a} and Matching-GAN \cite{DBLP:journals/corr/abs-2003-03497} combine the Matching Network \cite{DBLP:journals/corr/VinyalsBLKW16} (used for few-shot classification) with VAEs and GANs respectively. F2GAN \cite{DBLP:journals/corr/abs-2008-01999} enhances the fusion of high level features by filling low level details from the example images using a Non-local Attention module. Similarly, LoFGAN \cite{gu2021lofgan} is based upon a learnable Local feature Fusion module (LFM) combined with GANs. WaveGAN \cite{yang2022wavegan} adapts Haar Wavelet transform to capture features at different frequencies and fuses them. The images generated using these methods often lack in diversity due to their dependence on the few unseen examples available for the fusion operation.\\
\textbf{Optimization based approaches}, e.g.,\ FIGR \cite{https://doi.org/10.48550/arxiv.1901.02199} and DAWSON \cite{DBLP:journals/corr/abs-2001-00576} use adversarial learning (GANs) combined with meta-learning methods, e.g.,\ Reptile \cite{DBLP:journals/corr/abs-1803-02999} (used by \cite{https://doi.org/10.48550/arxiv.1901.02199}) and MAML \cite{DBLP:journals/corr/FinnAL17} (used by \cite{DBLP:journals/corr/abs-2001-00576}) to generate new images. LSO \cite{zheng2023my} adapts the latent space learnt using StyleGAN \cite{karras2020training} (an approach for training GANs with limited data) for each unseen class using optimization and semantic stabilization losses. Exposing the model \cite{zheng2023my} to the few unseen samples at test time helps it to capture the class-specific characteristics, allowing the generated samples to have high fidelity with the unseen class; however optimization needs to be performed carefully to ensure that the model does not over-fit on the few samples that can take away the generation quality and diversity.\\
\textbf{Transformation based approaches} (\eg, AGE \cite{ding2022attribute})learn the pattern of transformations between different pairs of the same class during training and use these transformations to generate new samples of novel categories from the available samples, e.g.,\ in DAGAN \cite{https://doi.org/10.48550/arxiv.1711.04340} any sample passed though the encoder gives a feature containing its class-level information. This feature along with a random vector is passed through the decoder to generate a different sample of the same class. DeltaGAN~\cite{HongDeltaGAN} learns to generate transformations called \textit{sample-specific deltas} which represent the information needed to convert the sample to another image of the same class. Given a sample, different plausible deltas can be generated based upon a random vector input.
However, the current transformation-based approaches require end-to-end training of transformation and generation that can be unstable resulting in generation of low quality images.

While Diffusion Models have been rarely explored for few-shot image synthesis, there are several works using diffusion models for few-shot generative \textit{adaptation}, i.e.,\ they pre-train the model using a large-scale dataset belonging to a source domain (that is related to the domain of the target few shot images), and then adapt it to the target image domain. For example, in D2C~\cite{DBLP:journals/corr/abs-2106-06819}, the first step involves joint training of a VAE and a Diffusion model in the VAE's latent space in an unsupervised manner on a large dataset. Then a classifier is trained over the learnt latent space of the VAE using the few labelled examples available. For instance, they use the CelebA-64 dataset \cite{DBLP:journals/corr/LiuLWT14} for the first step and then learn a binary classifier in the latent space to predict attributes like blonds/females using just 100 labelled examples. To generate images having these attributes, they first use the diffusion model to generate VAE latents from random noise. Then they pick those latents having high probability scores in the binary classifier and decode them through the VAE inference model.
Similarly, DDPM-PA~\cite{zhu2022few} devises a pairwise similarity loss to preserve the relative distances between generated samples during domain adaptation to achieve better sample diversity in the target domain.
To our knowledge, diffusion models for few-shot synthesis have only been investigated in FSDM~\cite{https://doi.org/10.48550/arxiv.2205.15463}, which uses a DDPM to generate images of different classes in a fusion approach-based manner. It obtains a set-level context by passing the set of images for each class along with the timestep embedding through a Vision Transformer~\cite{DBLP:journals/corr/abs-2010-11929} and uses the context to condition the generative path (reverse path) of the DDPM. However, its diffusion process lies in the image space which has the disadvantage of large memory requirements and longer inference time for high-resolution images. For this reason, this approach has been evaluated only on small-scale datasets having $32\times32$ images.
On the contrary, our approach ensures diversity and fidelity in the synthesized images by adequately modelling the distribution of the few shot samples in the conditional space of the diffusion process. This provides us with the flexibility to sample new and varied conditionals which are then effectively mapped to the image space. Since the diffusion itself happens in a regularized latent space, we are able to efficiently generate high-resolution and realistic novel class images.
\section{Method}
\label{sec:method}
\subsection{Preliminaries}
\textbf{Problem Definition} Given a dataset $\mathbb{D}=(\bx_j, y), y\in[1,\mathbb{C}], j\in[1,n_y]$ with $\mathbb{C}$ classes and $n_y$ images in each class, we divide the dataset into seen classes $\mathbb{C}_s$ and unseen classes $\mathbb{C}_u$ where $\mathbb{C}_s\cap\mathbb{C}_u=\phi$. Only the images from seen classes can be used while training. The task of $K-$ shot generation involves generating new images of any class from $\mathbb{C}_u$, using only $K$ images of this class.\\
\noindent Below, for the sake of completeness, we briefly revisit diffusion models \cite{DBLP:journals/corr/abs-2006-11239} and latent diffusion models \cite{rombach2021highresolution}, followed by the description of our proposed conditional distribution modelling approach for few-shot synthesis (\cref{section:Conditional_Space_Modelling})

\noindent \textbf{Diffusion Models} \cite{DBLP:journals/corr/abs-2006-11239} are a special class of generative models which configure the data distribution as a reverse (or \textit{denoising}) process of iteratively adding noise to the data.
Thus, the forward \textit{diffusion} process converts the structured data ($\mathbf{x}_0$) into pure noise ($\mathbf{x}_T$) in $T$ timesteps by adding small amounts of Gaussian noise each time. The amount of noise $\epsilon$ added at each step is controlled by a non-decreasing variance schedule $\left[\beta_t \in \left(0,1\right)\right]_{t=1}^{T}$. Thus at each step $t$, we sample $\epsilon\sim\mathcal{N}\left(0,I\right)$ and then get $\mathbf{x}_t=\sqrt{1-\beta_t}\bx_{t-1}+\sqrt{\beta_t}\epsilon$. After $T$ steps through the encoder, the information in the data point is destroyed completely and $\mathbf{x}_T$ becomes random Gaussian noise. 
The Denoising Diffusion Probabilistic Model (DDPM) \cite{DBLP:journals/corr/abs-2006-11239} learns the reverse, where we start from pure noise and recover a point from the original data distribution through $T$ steps of a decoder (denoted by $p_\theta$). Given the noisy data point $\mathbf{x}_t$ and the time step $t$ as input, the decoder predicts the noise $(\epsilon)$ added to the original data point that lead to this noisy input. The architecture most commonly used for $p_\theta$ is a time-conditioned UNet \cite{ronneberger2015u}. The training objective $(L_{vlb})$ comes by minimizing the variational bound on the negative log-likelihood of the data distribution and is used in combination with the mean square error $(L_{simple}= E_{t,x_0,\bm\epsilon}[\lVert\bm\epsilon - \bm\epsilon_{\theta}(\bx_t,t)\rVert^{2}])$ between the true noise and predicted noise. A detailed derivation of the training objective can be found in \cite{DBLP:journals/corr/abs-2006-11239}.

\noindent \textbf{Latent Diffusion Model} (LDM) \cite{rombach2021highresolution} provides an efficient setup for high resolution image synthesis. LDM benefits from the diversity and high sample quality offered by Diffusion models but does not trade off the computational efficiency by taking the diffusion process to the latent space of a pretrained Variational Autoencoder \cite{https://doi.org/10.48550/arxiv.1312.6114} (with Encoder $\mathcal{E}$ and Decoder $\mathcal{D}$). Since training the VAE beforehand leads to a compressed but meaningful latent space with only the imperceptible details gone, the latent diffusion process is able to model the semantic bits of the data much more efficiently as compared to the high dimensional image space. They also introduce the Cross-Attention mechanism \cite{vaswani2017attention} to effectively condition the diffusion on signals from other modalities. We adapt this conditional latent diffusion pipeline for our use case.
\begin{figure}
\begin{center}
\includegraphics[scale=0.24]{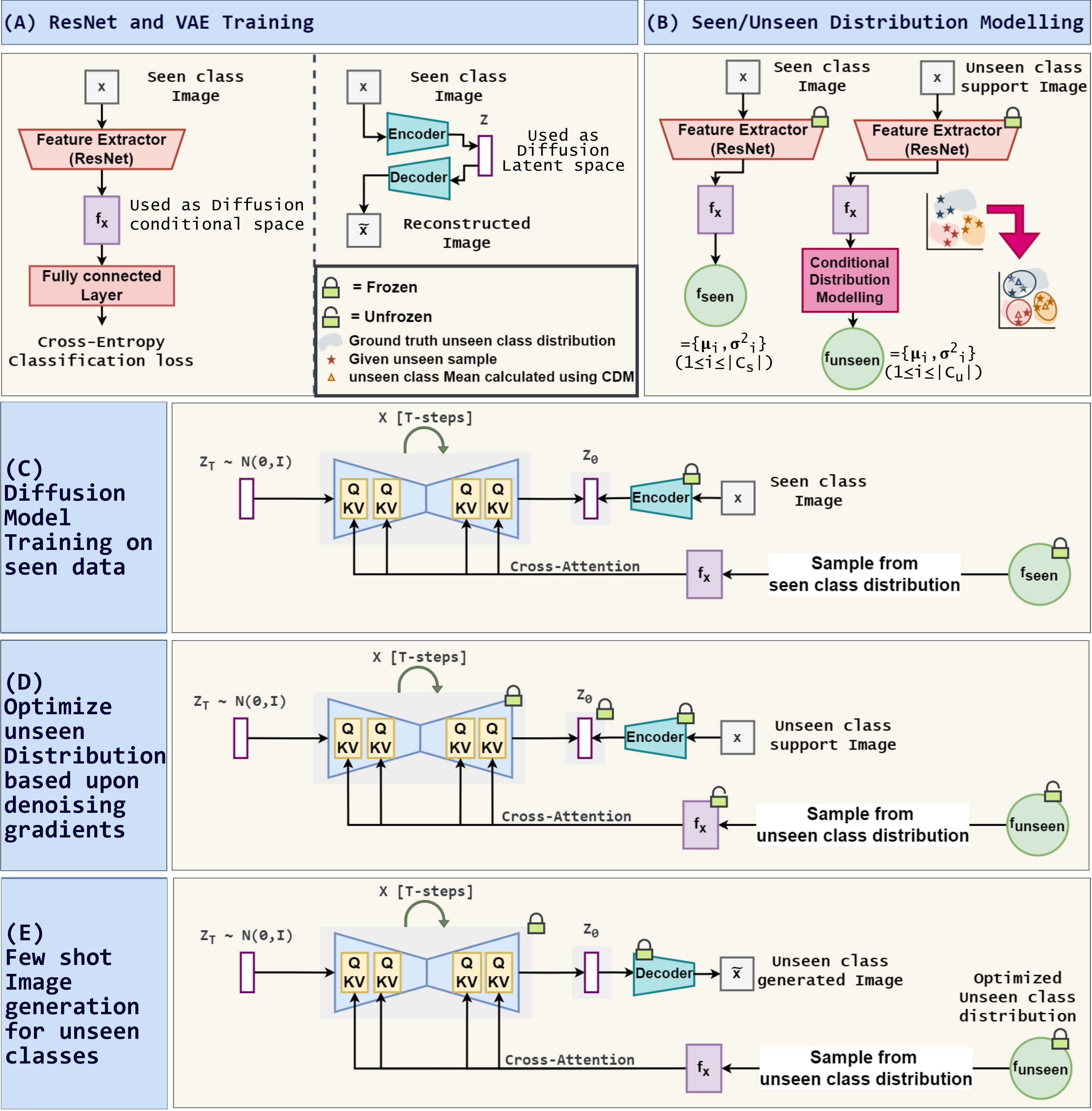}
\end{center}
\vspace{-1em}
   \caption{\textbf{CDM Pipeline: (A)} First, we train a ResNet classifier and a Variational Autoencoder (VAE) on the seen data (denoted by the set $\mathbb{C}_s$). The ResNet's penultimate layer's output ($f$) is to be used as a conditional input to the Diffusion model later. \textbf{(B)} We calculate the class-wise means ($\mu_i$) and variances ($\sigma^2_i$) of the seen data in the latent space $f$. This collection of Gaussian distributions is denoted by $f_{seen}$. Now, for each unseen class (belonging to the set $\mathbb{C}_u$), we have $K$ support samples for $K$-shot image generation task. The seen classes whose distributions ($\mu_i,\sigma^2_i$) are closest to this unseen class are used to estimate its distribution ($\mu_i,\sigma^2_i$) through Conditional Distribution Modelling. This process is shown in the two plots having 3 unseen classes, where 3 support samples per class and $f_{seen}$ are used approximate the unseen class distributions. \textbf{(C)} We train a Diffusion model in the VAE latent space, conditioned upon the samples obtained from the seen class distributions ($f_{seen}$). \textbf{(D)} We use inversion based optimization to improve the unseen class distributions ($f_{unseen}$) using the denoising gradients from the support samples. \textbf{(E)} The optimized unseen class distributions are used to generate new samples from the Diffusion Model.}
   \vspace{-1em}
\label{fig:CDM_idea}
\end{figure}
\subsection{Conditional Distribution Modelling (CDM)}
\label{section:Conditional_Space_Modelling}
As depicted in \cref{fig:CDM_idea}, Conditional Distribution Modelling involves five stages, each of which are described below.
\paragraph{A. ResNet and VAE Training.} (\cref{fig:CDM_idea} (A)) We initially train a vector-quantization \cite{van2017neural} regularized Variational Autoencoder (VAE) on the seen data (denoted by $\mathbb{C}_s$), following the LDM architecture. The LDM is trained at a later stage in this VAE's latent space, denoted by $\bz$. Simultaneously, we also train a simple ResNet-based classifier on the seen data using Cross-Entropy loss and choose the output of its penultimate layer (denoted by $f$) as the conditioning space for our LDM.
\paragraph{B. Distribution Modelling of seen and unseen classes.} (\cref{fig:CDM_idea} (B)) We obtain class-wise latent mean $\mu^y$ and latent variance ${\left(\sigma^2\right)}^y$ on the seen data as follows--
\begin{equation}
\label{Eq:Seen_means}
    \mu^y=\frac{1}{n_y}\sum_{j=1}^{n_y} f_j^y
\end{equation}
\begin{equation}
\label{Eq:Seen_variances}
    {\left(\sigma^2\right)}^y=\frac{1}{n_y-1}\sum_{j=1}^{n_y} \left(f_j^y-\mu^y\right)^2
\end{equation}
\begin{equation}
\label{Eq:Gaussian_distribution}
\mathbb{D}^y=\mathcal{N}\left({\mu}^y,{\left(\sigma^2\right)}^y\right)
\end{equation}
where $n_y$ is the number of samples in seen class $y$ and each dimension in the latent space $f$ is considered to be \textit{uncorrelated}. 
We assume that every dimension in the latent vector $f_j^y$ follows a Gaussian distribution (denoted by $\mathbb{D}_y$, $f_{seen}=\{\mathbb{D}_y\}_{y\in \mathbb{C}_s}$) and observe that similar classes usually have similar statistics (mean and variance) of the feature representations. This allows us to transfer the mean and variance statistics across similar classes, i.e.,\ from seen classes for which we have a better approximation of these statistics to the unseen classes for which we only have a few examples, which are insufficient to approximate the underlying distribution. 
\noindent Therefore, given $K$ support samples of an unseen class $c$($\in \mathbb{C}_u$), we first obtain the latent vectors $f^c_k$  ($1\leq k\leq K$) for each of these $K$ samples using the ResNet from stage (A). Then, we augment the latent space $f^c$ by calibrating the mean and variance statistics from the seen classes which are nearest to this unseen class. The nearest seen classes are the ones with the minimum Euclidean distance between their mean latent $\mu^y$ and the mean of the unseen latent $\mu^c$. The mean of an unseen class distribution is computed from latent vectors $f_k^c$ of support samples of that class: 
\begin{equation}
\label{Eq:Cal_mean}
\mu^c=\frac{\sum_{k=1}^{K}f_k^c}{K}
\end{equation}
Let the set of nearest seen classes be denoted by $\mathbb{S}_N$, the calibrated variance ${\left(\sigma^2\right)}^c$ of that unseen class distribution is given by: 
\begin{equation}
\label{Eq:Cal_variance}{\left(\sigma^2\right)}^c=\frac{\sum_{y\in\mathbb{S}_N}{\left(\sigma^2\right)}^y}{|\mathbb{S}_N|}
\end{equation}
\begin{algorithm}
\caption{Training LDM with CDM}\label{alg:CDM_training}
\begin{algorithmic}[1]
\Require Trained Feature Extractor and VAE
\Require Training data $D={(\bx_j, y)}, y\in\mathbb{C}_s, j\in[1,n_y]$
\Require Seen classes' statistics $\{\mu^y, \left(\sigma^2\right)^y\}, y\in \mathbb{C}_s$ obtained as per \cref{Eq:Seen_means} and \cref{Eq:Seen_variances}
\For{$m=1, ..., \#epochs$}
\For{$y=1, ..., |\mathbb{C}_s|$}
\For{$j=1, ..., n_y$}
\State Obtain the latent representation $\bz$ for $\bx_j$ by passing it through VAE Encoder
\State Sample a timestep $t\in[1,T]$ uniformly at random and get the noised version of $\bz$ denoted by $\bz_t$
\State Sample a latent $f$ for class $y$ from the gaussian distribution $\mathbb{D}^y$ as per \cref{Eq:Gaussian_distribution}
\State Use $f$ as a conditional input to the LDM to denoise $\bz_t$ into $\bz$ and update the UNet parameters based upon the loss $\mathcal{L}$ as per \cref{Eq:CDM_loss}.
\EndFor
\EndFor
\EndFor
\end{algorithmic}
\end{algorithm}
\paragraph{C. Diffusion Model Training.} (\cref{fig:CDM_idea} (C)) In this stage, we train a Latent Diffusion Model (LDM) on the seen data in the $z$-space of the stage (A) VAE. This LDM is conditioned using samples from the seen data distributions $f_{seen}$ defined in (B). Thus, given an image $\bx$ belonging to a seen class $y$, we first sample a latent $f^y \sim \mathbb{D}^y$ (the corresponding distribution obtained as per \cref{Eq:Gaussian_distribution}) and also get the perceptually compressed representation $\bz$ corresponding to $\bx$ from the pretrained VAE. Now, the LDM learns to denoise the noisy version  $\bz_t$ of $\bz$, conditioned on $f^y$ and time-stamp $t$ which can be shown as the overall training objective \cite{DBLP:journals/corr/abs-2102-09672},
\begin{equation}
\label{Eq:CDM_loss}
    \mathcal{L} := \mathcal{L}_{simple}+\lambda \mathcal{L}_{vlb},
\end{equation}
where $\mathcal{L}_{vlb}$ is the variational lower bound based loss and $\mathcal{L}_{simple}$ is the mean squared error between the true noise and predicted noise and is given by
\begin{equation}
    \mathcal{L}_{simple} := \mathbb{E}_{\mathcal{E}(\bx),y,\epsilon\sim\mathcal{N}\left(0,I\right),t}\left[\lVert\bm\epsilon - \bm\epsilon_{\theta}(\bz_t,t,f^y)\rVert^{2}_{2}\right]
\end{equation}

\noindent We further employ classifier-free guidance proposed by \cite{https://doi.org/10.48550/arxiv.2207.12598} in order to enhance the sample diversity. In summary, while training the LDM, we ensure that the model learns to \textit{transform} the vectors $f^y$ that are sampled from the class distribution $\mathbb{D}^y$ into the latent representations $\bz$ of the same class. Algorithm \ref{alg:CDM_training} describes the LDM training process with the proposed CDM.
\begin{algorithm}
\caption{Inversion based optimization of unseen class distributions}\label{alg:CDM_inversion}
\begin{algorithmic}[1]
\Require $K$ unseen class samples $\{\bx_i,c\}_{i=1}^K, c\in \mathbb{C}_u$
\Require Trained VAE and LDM models
\Require Initial unseen classes' statistics $\{\mu^c,\left(\sigma^2\right)^c\}, c\in \mathbb{C}_u$ obtained as per \cref{Eq:Cal_mean} and \cref{Eq:Cal_variance}
\For{$m=1, ..., \#optimization\;steps$}
\For{$c=1, ..., |\mathbb{C}_u|$}
\For{$i=1, ..., K$}
\State Obtain the latent representation $\bz$ for $\bx_i$ by passing it through VAE Encoder
\State Sample a timestep $t\in[1,T]$ uniformly at random and get the noised version of $\bz$ denoted by $\bz_t$
\State Sample a latent $f^c$ for class $c$ by first sampling a latent $\bm\epsilon$ from the standard normal distribution $\bm\epsilon\sim\mathcal{N}(0,I)$ and then getting $f^c=\mu^c+\sigma^c*\bm\epsilon$
\State Use $f^c$ as conditional input to the LDM to denoise $\bz_t$ into $\bz$ and update $\mu^c,\left(\sigma^2\right)^c$
\EndFor
\EndFor
\EndFor
\end{algorithmic}
\end{algorithm}
\paragraph{D. Inversion based optimization of unseen class distributions.} (\cref{fig:CDM_idea} (D)) Once the diffusion model is trained, we can use the unseen class distributions defined in stage (B) (denoted by $f_{unseen}$) to generate unseen class samples from the model, however, we observe that the samples generated using these distributions are more similar to the seen classes which are in the neighbourhood of the unseen classes. We hypothesize that the few conditionals $f^c_k$ corresponding to the support samples of unseen class $c$ are not able to properly capture the characteristics of the unseen class and hence, the unseen class distribution ends up being very close to the neighbouring seen class distributions. Hence, we propose to refine the unseen distributions using \textit{inversion} based optimization, i.e., given a support image $\bs$ from an unseen class $c$, we aim to find the conditional $f^c_{\bs}$ that results in the construction of $\bs$ from our frozen LDM. Since the conditional $f^c_{\bs}$ is sampled from the Gaussian distribution of class $c$, we back-propagate the gradients to optimize the mean $\mu^c$ and variance $\left(\sigma^2\right)^c$ as well. Our optimization goal can therefore be defined as
\begin{equation}
\label{eq:Inversion_optimization}
\left(\mu^c\right)^*, \left(\left(\sigma^2\right)^c\right)^* = \underset{\mu^c, \left(\sigma^2\right)^c}{\arg\!\min}\;\mathbb{E}_{\mathcal{E}(\bs),c,\epsilon\sim\mathcal{N}\left(0,I\right),t}\left[\lVert\bm\epsilon - \bm\epsilon_{\theta}(\bz_t,t,f^c_{\bs})\rVert^{2}_{2}\right]
\end{equation}
Algorithm \ref{alg:CDM_inversion} describes the inversion based optimization of unseen class distributions.
\paragraph{E. Few-shot Image generation using optimized unseen distributions.} (\cref{fig:CDM_idea} (E)) Once our unseen class distributions are optimized, we can generate diverse samples for each class using the LDM by sampling $\bz_T\sim\mathcal{N}(0,I)$ and $f^c\sim\mathcal{N}\left(\mu^c,\left(\sigma^2\right)^c\right)$.

\begin{table*}[h]
\begin{center}
\begin{adjustbox}{max width=\textwidth}
\begin{tabular}{l|c|cc|cc|cc|cc}
\toprule
\multirow{2}{*}{Method} & \multirow{2}{*}{Shot} & \multicolumn{2}{c|}{Flowers} & \multicolumn{2}{c|}{Animal Faces}& \multicolumn{2}{c|}{VGGFace} & \multicolumn{2}{c}{NABirds}\\
& & FID$(\downarrow)$ & LPIPS$(\uparrow)$ & FID$(\downarrow)$ & LPIPS$(\uparrow)$& FID$(\downarrow)$ & LPIPS$(\uparrow)$& FID$(\downarrow)$ & LPIPS$(\uparrow)$\\
\midrule
\multirow{2}{*}{DAGAN\cite{https://doi.org/10.48550/arxiv.1711.04340}} & 3 & 151.21 & 0.0812 & 155.29 & 0.0892& 128.34 & 0.0913 & 159.69 & 0.1405\\
\cline{2-10}
 & 1 & \textit{179.59} & \textit{0.0496} & \textit{185.54} & \textit{0.0687} & \textit{134.28} & \textit{0.0608} & \textit{183.57} & \textit{0.0967}\\
\hline
MatchingGAN\cite{DBLP:journals/corr/abs-2003-03497} & 3 & 143.35 & 0.1627 & 148.52 & 0.1514& 118.62 & 0.1695 & 142.52 & 0.1915\\
\hline
F2GAN\cite{DBLP:journals/corr/abs-2008-01999} & 3 & 120.48 & 0.2172 & 117.74 & 0.1831 & 109.16 & 0.2125 & 126.15 & 0.2015\\
\hline
LoFGAN\cite{gu2021lofgan} & 3 & 112.55 & 0.2687 & 116.45 & 0.1756& 106.24 & 0.2096 & 124.56 & 0.2041\\
\hline
\multirow{2}{*}{DeltaGAN\cite{HongDeltaGAN}} & 3 & 104.62 & \textbf{0.4281} & 87.04 & \textbf{0.4642}& 78.35 & \textbf{0.3487} & 95.97 & 0.5136\\
\cline{2-10}
& 1 & \textit{109.78} & \textit{\textbf{0.3912}} & \textit{89.81} & \textit{\textbf{0.4418}}& \textit{80.12} & \textit{\textbf{0.3146}} & \textit{96.79} & \textit{\textbf{0.5069}}\\
\hline
\multirow{2}{*}{LSO\cite{zheng2023my}} & 3 & \textbf{47.34} & 0.3805 & \underline{43.29} & 0.4446& \textbf{4.77} & \underline{0.2835} & \textbf{21.67} & \underline{0.5347}\\
\cline{2-10}
& 1  & \textit{\textbf{55.79}} & \textit{0.2721} & \textit{\textbf{64.35}} & \textit{0.2230} & \textit{\textbf{5.88}} & \textit{\underline{0.1650}} & \textit{\textbf{25.23}} & \textit{0.3318}\\
\hline
\multirow{2}{*}{CDM (Ours)} & 3 & \underline{77.26} & \underline{0.4034} & \textbf{40.04} & \underline{0.4459}& \underline{12.77} & 0.2029 & \underline{41.48} & \textbf{0.5406}\\
\cline{2-10}
& 1 &  \textit{\underline{83.97}} & \textit{\underline{0.3728}} & \textit{\underline{64.88}} & \textit{\underline{0.3077}}&  \textit{\underline{12.61}} & \textit{0.1240} & \textit{\underline{45.53}} & \textit{\underline{0.4958}}\\
\bottomrule
\end{tabular}
\end{adjustbox}
\end{center}
\caption{FID $(\downarrow)$ and LPIPS $(\uparrow)$ of images generated by different methods for unseen categories on four datasets in 1 and 3-shot setting. The best results for each shot are in bold and the second best results have been underlined. 1-shot results are displayed in italics. We quote the results of the baseline methods from the DeltaGAN paper\cite{HongDeltaGAN}. The LSO\cite{zheng2023my} results are obtained by running their official code using our setting.}
\label{Table:Quantitative_Results}
\vspace{-2em}
\end{table*}

\section{Experiments}
\label{sec:experiments}
\subsection{Evaluation Setup}
For our evaluation, we follow the setup in recent state-of-the-art approach \cite{HongDeltaGAN}. Given a dataset $\mathbb{D}=(\bx_j, y), y\in[1,\mathbb{C}], j\in[1,n_y]$ with $\mathbb{C}$ classes and $n_y$ images in each class, we split it into $\mathbb{C}_u$ unseen and $\mathbb{C}_s$ seen categories. After training on $\mathbb{C}_s$, at test time for $K$-shot generation, the model uses $K$ images and generates 128 fake images of each unseen class giving a set $S_{fake}$ of $128|\mathbb{C}_u|$ fake images. These fake images are generated by considering 2 nearest classes corresponding to each image for conditional space modelling (stage (B) in \cref{sec:method}) and using 25 steps of DDIM sampling \cite{DBLP:journals/corr/abs-2010-02502} in the diffusion model. The remaining $n_y-K$ images are combined from each unseen category to obtain a set $S_{real}$ of real images. For quantitative comparisons, we calculate the Fréchet Inception Distance (FID)~\cite{10.5555/3295222.3295408} between $S_{real}$ and $S_{fake}$ and the Learned Perceptual Image Patch Similarity (LPIPS)~\cite{8578166} for $S_{fake}$. FID score measures the distance between the latent features of real unseen images and generated unseen images and is an indicator of image fidelity. The latent space used to calculate the distance is from an Inception-V3 \cite{43022} architecture pretrained on the Imagenet dataset \cite{deng2009imagenet}. LPIPS indicates the diversity in the generated images by calculating the distance between all the same class image pairs and then averaging it for each class followed by measuring the average over all the unseen classes.
We use the following four datasets in order to compare our method with the existing approaches:\\
\noindent \textbf{Flowers} \cite{nilsback2008automated} dataset has total 102 categories with number of images in each class varying from 40 to 258. We split it into 85 seen and 17 unseen classes resulting in 7121 seen and 680 unseen images.\\
\noindent \textbf{Animal Faces} \cite{deng2009imagenet} dataset contains 149 classes. We split it into 119 seen categories having total 96621 images and 30 unseen categories with 3000 images.\\
\noindent In \textbf{VGGFace} \cite{cao2018vggface2} dataset, we select 2299 classes and choose 100 images for each class. 1802 classes out of these are used for training and the remaining 497 for evaluation.\\
\noindent \textbf{NABirds} \cite{van2015building} dataset has 555 classes out of which 444 are used for training and 111 for evaluation. This gives a total of 38306 seen images and 10221 unseen images.

\subsection{Implementation Details}
For the Variational Autoencoder (VAE), we set the downsampling factor to 4 in all our experiments, i.e.\ given the image-space $\bx\in\mathbb{R}^{128\times128\times3}$, the latent space $\bz\in\mathbb{R}^{32\times32\times3}$. We maintain a codebook $\mathbb{Z}$ of size 2048 for the vector-quantization based regularization and set the batch size to 16 for all other datasets except VGGFace (due its large size), for which we set it to 128. We train the VAE for 100 epochs. For the conditional space, we train ResNet-50 \cite{DBLP:journals/corr/HeZRS15} using Cross-Entropy Loss and SGD optimizer for 500 epochs on seen data. Our conditional latent space is then constructed by taking the output of the last latent convolution layer having dimensions $4\times4\times512$. Finally, the conditional latent diffusion pipeline is learnt which uses 1000 timesteps and a linear noise schedule. The value of $\lambda$ in the training objective \cref{Eq:CDM_loss} is set to $0.001$ following the previous works \cite{DBLP:journals/corr/abs-2102-09672}. We use a batch size of 128 to train the LDM for around 200 epochs. The learning rate has a warmup schedule, increasing linearly from $0$ to $2e-4$ in 50K steps. It takes roughly 24 hours to train the diffusion model on a single NVIDIA GeForce RTX 3060 GPU for all the datasets except VGGFace. Due to its large size, the LDM for VGGFace is trained on a single Nvidia RTX A6000 GPU for nearly 70 hours. The inversion based optimization is performed with a learning rate of $2e-4$ for 50K steps.


\subsection{Quantitative Evaluation}
We compare our proposed approach against several competing methods. These include DAGAN \cite{https://doi.org/10.48550/arxiv.1711.04340}, MatchingGAN \cite{DBLP:journals/corr/abs-2003-03497}, F2GAN \cite{DBLP:journals/corr/abs-2008-01999}, LoFGAN \cite{gu2021lofgan} DeltaGAN \cite{HongDeltaGAN} and LSO \cite{zheng2023my} in both 3 and 1 shot settings. For fusion-based approaches, we report only 3-shot results, since they can not be evaluated in 1 shot setting. For the transformation-based approaches, which need only one conditional image, 3-shot setting implies that $S_{fake}$ is generated using 3 images per episode, choosing one randomly out of the 3 each time. The results are summarized in \cref{Table:Quantitative_Results} on two datasets. The results shows that our method performs comparable to GAN-based approaches in terms of both sample quality as well as diversity. For example, for the Flowers dataset, our model is second-best in terms of both FID (sample quality) as well as LPIPS (sample diversity) in both 1 and 3 shot settings. While LSO obtains the best sample quality, the generated samples lack diversity. Similarly, DeltaGAN samples, though diverse, lack in terms of quality. On the AnimalFaces dataset, our approach achieves the best FID score and second-best LPIPS score in 3-shot setting.
\subsubsection{Few-Shot Classification}
To further evaluate the representativeness of the generated samples, we present the results of Few-shot classification performed on a held-out test set comprising of unseen classes. For $N$-way $C$-shot classification, we randomly choose $N$ unseen classes and use $C$ images of each class to generate $512$ fake images. A ResNet18 pretrained on seen data is used as a feature extractor and the final linear layer is trained using $N\times(C+512)$ images. The rest of the unseen images are used for testing. As \cref{Table:FS_classification_results} shows, the models trained using CDM-generated samples are able to classify the test set images with performance comparable to state-of-the-art GAN-based methods. Therefore, the generated samples match the characteristics of the unseen classes.
\begin{table}[!t]
\noindent\begin{minipage}[!t]{.5\linewidth}
\centering
\begin{tabular}{l|cc}
\hline
Method & 1-shot & 5-shot\\
\hline
MatchingGAN\cite{DBLP:journals/corr/abs-2003-03497} & - & 74.09\\
LoFGAN\cite{gu2021lofgan} & - & 75.86\\
DeltaGAN\cite{HongDeltaGAN} & \textbf{61.23} & 77.09\\
LSO\cite{zheng2023my}&57.42&\textbf{79.41}\\
CDM (Ours) & \underline{60.12} & \underline{78.99}\\
\hline
\end{tabular}
\vspace{1em}
\captionof{table}{Accuracy(\%) of different methods on Flowers dataset in few-shot classification setting (10-way 1/5-shot) averaged over 10 episodes. Results of prior methods are as per DeltaGAN \cite{HongDeltaGAN}. The results of LSO \cite{zheng2023my} have been calculated using their official code in our setting.}
\label{Table:FS_classification_results}
\end{minipage}\hspace{1em}
\begin{minipage}[!t]{.5\linewidth}
\centering
\begin{tabular}{l|c|c}
\hline
Score & \ding{55} Inversion & \ding{51} Inversion\\
\hline
FID$(\downarrow)$ & 85.42 & 40.04 \\
LPIPS$(\uparrow)$ & 0.5684 & 0.4459 \\
\hline
\end{tabular}
\vspace{1em}
\captionof{table}{3-shot FID$(\downarrow)$ and LPIPS$(\uparrow)$ scores on AnimalFaces dataset, when applying CDM with and without inversion based optimization of the unseen class distributions.}
\label{Table:InversionEffect_Animalfaces}
\end{minipage}
\vspace{-2em}
\end{table}
\begin{figure*}[!t]
\begin{center}
\includegraphics[scale=0.32]{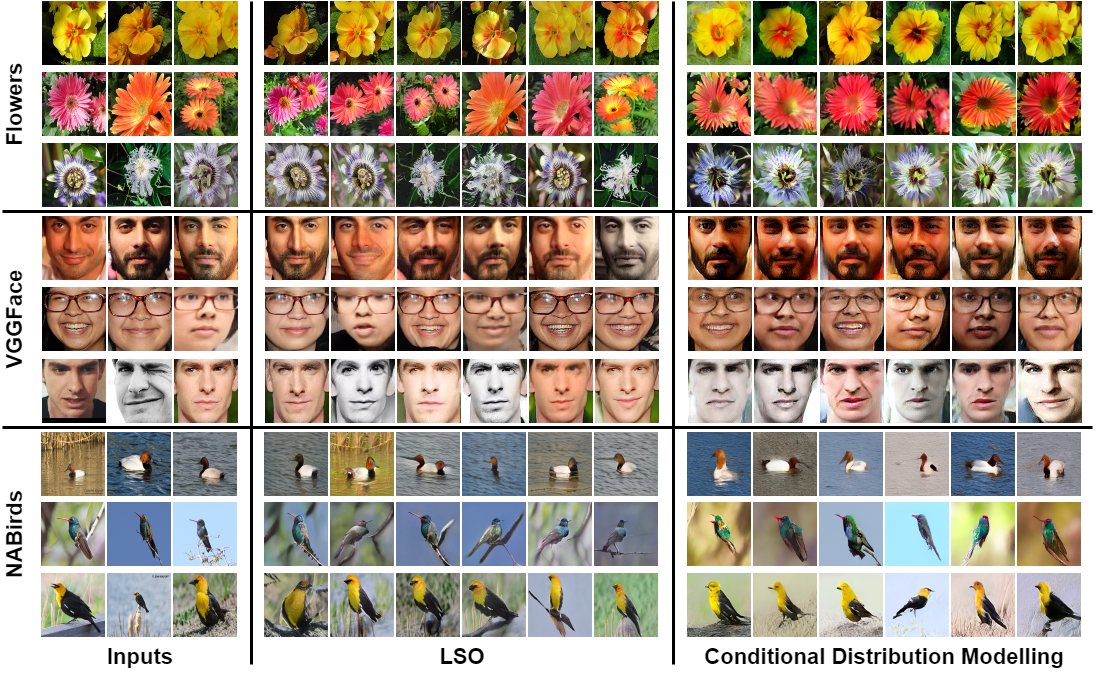}
\end{center}
\vspace{-1em}
   \caption{Comparison based on images generated by LSO \cite{zheng2023my} and our Conditional Distribution Modelling in 3-shot setting on Flowers, VGGFace and NABirds datasets. The conditional images are in the input columns.}
\label{fig:3-shot_visualize}
\vspace{-1em}
\end{figure*}
\begin{figure*}[!t]
\begin{center}
\includegraphics[scale=0.38]{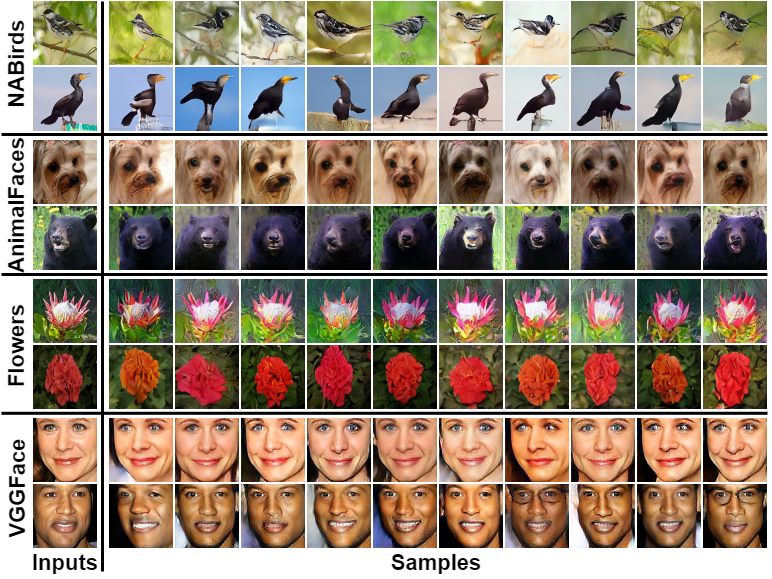}
\end{center}
\vspace{-1em}
   \caption{Samples generated in 1-shot setting using CDM on four datasets. The conditional images are in the input column.}
\label{fig:1-shot_visualize}
\vspace{-2em}
\end{figure*}
\subsection{Qualitative Results}
For qualitative comparison, we show some generated samples using LSO \cite{zheng2023my} in 3-shot setting in \cref{fig:3-shot_visualize} on Flowers, VGGFace and NABirds datasets. For all the datasets, the images generated by CDM are comparable to the ones produced by the state-of-the-art LSO in terms of quality as well as diversity. We observe changes in pose/orientation and colour transfers when compared to the conditional (support) images for both CDM and LSO. Rather, for VGGFace dataset, we can observe a doppelganger kind of effect on the generated samples for LSO, i.e., the generated faces don't necessarily have the same identity as the input faces (evident in rows 4 and 6). This is not the case with CDM, where the generated faces are considerably diverse, yet preserve the input identity. The same effect is visible in the NABirds dataset samples too, in row 8 (column 5) and row 9 (column 4), where the generated bird appears to be from a different (yet considerably similar) category than the input images, which does not happen with CDM. Perhaps the CDM samples lag in terms of FID scores, primarily because of poorer background generation as compared to LSO, and not due to the quality of birds themselves.\\
In \cref{fig:1-shot_visualize}, we show some samples generated in 1-shot setting on the NABirds, AnimalFaces, Flowers and VGGFace datasets using our approach. The generated samples display a wide range of poses/orientations/expressions, while maintaining the definitive characteristics of the input sample for all four datasets.
\subsection{Ablation Experiments}
\begin{figure*}[!b]
\begin{center}
\includegraphics[scale=0.32]{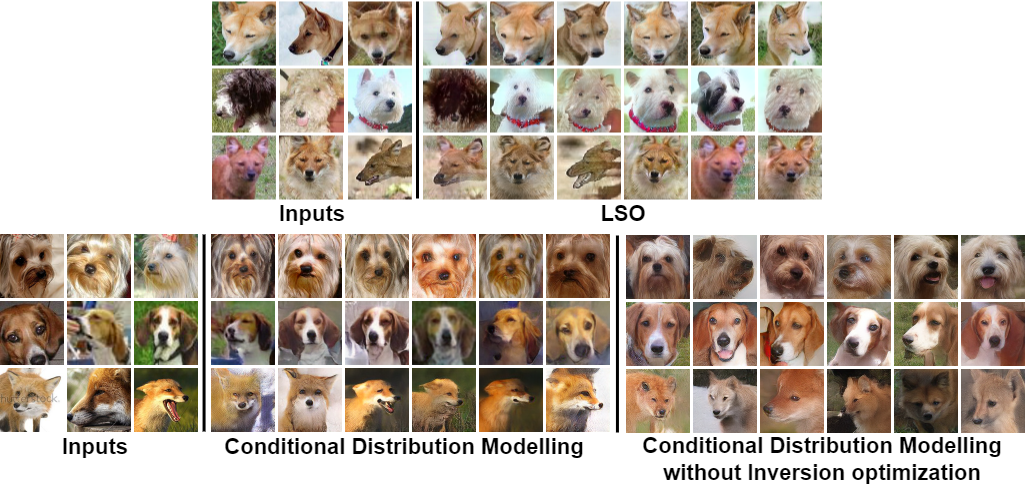}
\end{center}
\vspace{-1em}
   \caption{Comparison based on images generated by LSO \cite{zheng2023my} and our Conditional Distribution Modelling in 3-shot setting on AnimalFaces dataset. The conditional images are in the input columns. We also show the images generated by CDM with and without inversion based optimization for comparison.}
\label{fig:3-shot_visualize_animalfaces}
\end{figure*}
\noindent \textbf{Effect of inversion based optimization} In order to demonstrate the effectiveness of optimizing the unseen class distributions using inversion \cref{section:Conditional_Space_Modelling}(D), we calculate the FID and LPIPS scores of the samples generated with and without inversion based optimization on the AnimalFaces dataset in 3-shot setting. As observed in \cref{Table:InversionEffect_Animalfaces}, the image quality improves drastically using inversion based optimization, although the diversity of the generated samples drops. The improvement in the generation quality can be attributed to the increased fidelity of the samples after optimization; when inversion is not applied, samples belonging to neighbouring seen classes are generated. This is also evident in \cref{fig:3-shot_visualize_animalfaces} -- for a specific species of \textit{fox} in AnimalFaces, CDM without inversion tends to produce a different species of \textit{fox}, one that the model has seen while training. This is not the case when inversion is applied.

\section{Conclusion and Future Directions}
In this work, we have proposed \textit{Conditional Distribution Modelling (CDM)} - a framework that successfully employs Diffusion models for few-shot image synthesis on large scale datasets and achieves state-of-the-art results. We show how taking advantage of the neighbouring seen class statistics at test time can greatly benefit the image generation diversity. However, like all the diffusion based approaches, its generation time is longer than GAN-based works because of the multiple forward passes required during the denoising process. 

%
%
\bibliographystyle{splncs04}
\bibliography{main}
\end{document}